\documentclass[11pt]{article}

\usepackage[preprint]{acl}

\usepackage{times}
\usepackage{latexsym}

\usepackage[T1]{fontenc}

\usepackage[utf8]{inputenc}

\usepackage{microtype}

\usepackage{inconsolata}

\usepackage{graphicx}
\usepackage{multirow}
\usepackage{xurl}
\usepackage[english]{babel}
\usepackage{booktabs}    
\usepackage{xcolor}      
\usepackage{colortbl}    
\usepackage{tabularx}    
\usepackage{graphicx}    
\usepackage{subcaption}  
\usepackage{amsmath}     
\usepackage{arydshln}    
\usepackage{cleveref}
\usepackage{amssymb}
\usepackage{amsmath}
\usepackage{fontawesome}

\definecolor{tableheader}{HTML}{ffffff}
\definecolor{groupheader}{HTML}{eeeeee}
\definecolor{deltapos}{HTML}{EF4444} 
\definecolor{deltaneg}{HTML}{10B981} 

\definecolor{deltaneu}{HTML}{94A3B8} 

\definecolor{baselinebg}{HTML}{ffffff}
\definecolor{block2bg}{HTML}{DFEEFF}
\definecolor{block3bg}{HTML}{ECFDF5}

\newcommand{\deltafont}[2]{{\scriptsize\color{#1}(#2)}}
\newcommand{\best}[1]{\textbf{#1}}
\newcommand{\second}[1]{\underline{#1}}
\definecolor{revcolor}{HTML}{000000} 
\newcommand{\rev}[1]{\textcolor{revcolor}{#1}}

\title{OGER: A Robust Offline-Guided Exploration Reward for Hybrid Reinforcement Learning}

\author{
  \textbf{Xinyu Ma\textsuperscript{1,*}},
  \textbf{Mingzhou Xu\textsuperscript{2,*}},
  \textbf{Xuebo Liu\textsuperscript{1,$\dagger$}},
  \textbf{Chang Jin\textsuperscript{2}},
\\
  \textbf{Qiang Wang\textsuperscript{2}},
  \textbf{Derek F. Wong\textsuperscript{3}},
  \textbf{Min Zhang\textsuperscript{1}}
\\
\\
  \textsuperscript{1}Institute of Computing and Intelligence, Harbin Institute of Technology, Shenzhen, China \\
  \textsuperscript{2}Hithink RoyalFlush Information Network, Hangzhou, China \\
  \textsuperscript{3}Computer and Information Science, University of Macau, Macau, China
\\
\textsuperscript{*}Equal contribution.
\textsuperscript{$\dagger$}Corresponding author.
\\
  \small{
    \faEnvelopeO~: \href{mailto:liuxuebo@hit.edu.cn}{liuxuebo@hit.edu.cn}
  }
}

\begin{document}
\maketitle
\begin{abstract}
Recent advancements in Reinforcement Learning with Verifiable Rewards (RLVR) have significantly improved Large Language Model (LLM) reasoning, yet models often struggle to explore novel trajectories beyond their initial policy distribution. While offline teacher guidance and entropy-driven strategies have been proposed to address this, they often lack deep integration or are constrained by the model's inherent capacity.
In this paper, we propose OGER (\second{\textbf{O}}ffline-\second{\textbf{G}}uided \second{\textbf{E}}xploration \second{\textbf{R}}eward), a novel framework that unifies offline teacher guidance and online reinforcement learning through a specialized reward modeling lens. OGER employs multi-teacher collaborative training and constructs an auxiliary exploration reward that leverages both offline trajectories and the model's own entropy to incentivize autonomous exploration. 
Extensive experiments across mathematical and general reasoning benchmarks demonstrate that OGER consistently outperforms competitive baselines, achieving substantial gains in mathematical reasoning while maintaining robust generalization to out-of-domain tasks. We provide a comprehensive analysis of training dynamics and conduct detailed ablation studies to validate the effectiveness of our entropy-aware reward modulation. 
Our code is available at \url{https://github.com/ecoli-hit/OGER.git}.
\end{abstract}

\section{Introduction}
Current Large Language Models (LLMs) have demonstrated increasingly sophisticated reasoning capabilities~\citep{liu2024deepseek,yang2025qwen3}. 
This progress is primarily catalyzed by Reinforcement Learning with Verifiable Rewards (RLVR), which fosters the emergence of extended chains of thought and strengthens systemic reasoning. Following the paradigm established by DeepSeek-R1~\citep{guo2025deepseek}, subsequent advancements~\citep{du2025confidencerewardtransformingllms,feng2025don,liu2025attentioncompassefficientexploration,wang2025aspoasymmetricimportancesampling,deng2026rearl} have focused on augmenting model performance by refining the Group Relative Policy Optimization (GRPO)~\citep{shao2024deepseekmath} algorithm. 

Despite these strides, recent findings~\citep{yue2025does,nguyen2025reasoningboundaryparadoxreinforcement} consistently highlight a critical bottleneck: while RLVR significantly elevates the performance of the foundational model, it often fails to foster truly novel problem-solving capabilities. \citet{zhao2025echo} identify an ``echo chamber'' effect, wherein reinforcement learning (RL) converges toward dominant pre-existing distributions. Similarly, \citet{nguyen2025reasoningboundaryparadoxreinforcement, yue2025does} argue that self-sampling primarily reinforces known correct patterns rather than catalyzing the discovery of original reasoning trajectories.
To address these limitations, research has branched into two primary directions: incorporating offline guidance and applying entropy-based regularization. On one hand, frameworks such as Luffy~\citep{luffy} and Chord~\citep{zhang2025onpolicyrlmeetsoffpolicy} leverage high-quality teacher trajectories to provide gold-standard demonstrations, facilitating a rapid transition from supervised imitation to autonomous reasoning. On the other hand, entropy-driven approaches~\citep{cui2025entropymechanismreinforcementlearning, wang20258020rulehighentropyminority} mitigate ``entropy collapse'' by maintaining the model within a high-entropy regime. By sustaining diverse exploration and preventing premature convergence to suboptimal, these methods ensure the continued discovery of effective paths.

However, while external teacher data provides an intuitive performance boost, existing methods often lack a seamless integration between offline and online components. Furthermore, entropy-driven strategies remain fundamentally constrained by the model's inherent capacity. To bridge this gap, we propose \textbf{OGER}, a framework that enhances reasoning capabilities by unifying offline teacher guidance and online RL through a specialized reward modeling lens. 
Specifically, our framework leverages multi-source teacher trajectories for collaborative training to solidify foundational reasoning capabilities. Concurrently, we construct a divergence-based exploration reward to quantify the semantic disparity between online and offline trajectories, facilitating a profound synergy between expert imitation and autonomous discovery. To ensure training stability, we implement a hybrid sampling mechanism that integrates offline expert data directly into the online training batches. Furthermore, we refine this exploration signal by incorporating the policy model's token-level entropy distribution, enabling fine-grained control to incentivize novel reasoning behaviors and mitigate the risk of premature convergence.
Our main contributions are as follows:
\begin{itemize}
    \item We propose OGER, a sophisticated RLVR framework that integrates multi-teacher offline trajectories into RL. By introducing a novel offline-guided exploration reward and leveraging entropy for reward shaping, OGER effectively harmonizes offline knowledge with online discovery.
    \item We demonstrate that OGER functions as a high-quality reward mechanism, achieving significant performance gains over competitive baselines, including 4.3\% and 6.9\% improvement on average of mathematics and general evaluation with 1.5B and 7B models.
    \item We provide in-depth analysis of our OGER framework’s efficacy, analyzing the evolution of scores and entropy during training, and conducting comprehensive ablation studies of its modules to validate individual contributions.
\end{itemize}

\section{Related Work}

\subsection{Off-policy Guided Reinforcement Learning}
Integrating off-policy data into RLVR has emerged as a promising frontier for augmenting LLMs' reasoning capabilities, leading to several distinct research trajectories. One approach emphasizes the collection, filtration, and replay of historical trajectories generated by the online model~\citep{zhan2025exgrpolearningreasonexperience,dou2025improvingrlexplorationllm}, aiming to maximize sample efficiency and exploit the model's past experiences. Another line of inquiry investigates the fusion of Supervised Fine-Tuning (SFT) and RL, where the relative weighting of losses is dynamically adjusted to balance imitation and exploration~\citep{lanchantin2025bridgingofflineonlinereinforcement,lv2025unifiedviewlargelanguage}. More recently, advancements have focused on the adaptive mixing of offline teacher data with online samples, utilizing importance sampling and sophisticated weighting schemes to shift the model's focus during training~\citep{luffy, zhang2025onpolicyrlmeetsoffpolicy, bartoldson2025trajectorybalanceasynchronydecoupling}.


\subsection{The Entropy Mechanism in RLVR}
Despite the success of RLVR, recent studies~\citep{yue2025does,karan2025reasoningsamplingbasemodel,xie2025unlockingexplorationrlvruncertaintyaware,ke-etal-2025-aquilt,deng-etal-2025-drpruning} suggest that RL primarily amplifies pre-existing capabilities within the base model rather than inducing entirely novel reasoning paradigms. \citet{cui2025entropymechanismreinforcementlearning} analyzes this phenomenon through the lens of information theory, identifying the challenge of entropy collapse: without robust regularization, policy entropy diminishes precipitously during early training, leading the model to converge prematurely to a narrow subset of high-reward trajectories. Consequently, sustaining a relatively high-entropy regime is recognized as a pivotal factor in preventing entropy collapse and facilitating more profound reasoning discovery.

Current research has explored various entropy-centric methodologies to revitalize training. For instance, \citet{wang20258020rulehighentropyminority} demonstrates that selectively training on the top 20\% of tokens with the highest entropy outperforms standard full-token fine-tuning. \citet{cheng2025reasoningexplorationentropyperspective} introduces entropy-based advantage shaping to bolster exploratory reasoning, while \citet{zhang2025rediscoveringentropyregularizationadaptive} proposes an adaptive regularization framework to mitigate entropy decay throughout the training lifecycle. Furthermore, recent works~\citep{su2025cegppocontrollingentropygradientpreserving,wang2025arbitraryentropypolicyoptimization} leverage entropy for granular analysis of model generations to achieve a more nuanced balance between exploration and exploitation.


\begin{figure*}[!tbp]
    \centering
    \includegraphics[width=0.85\linewidth]{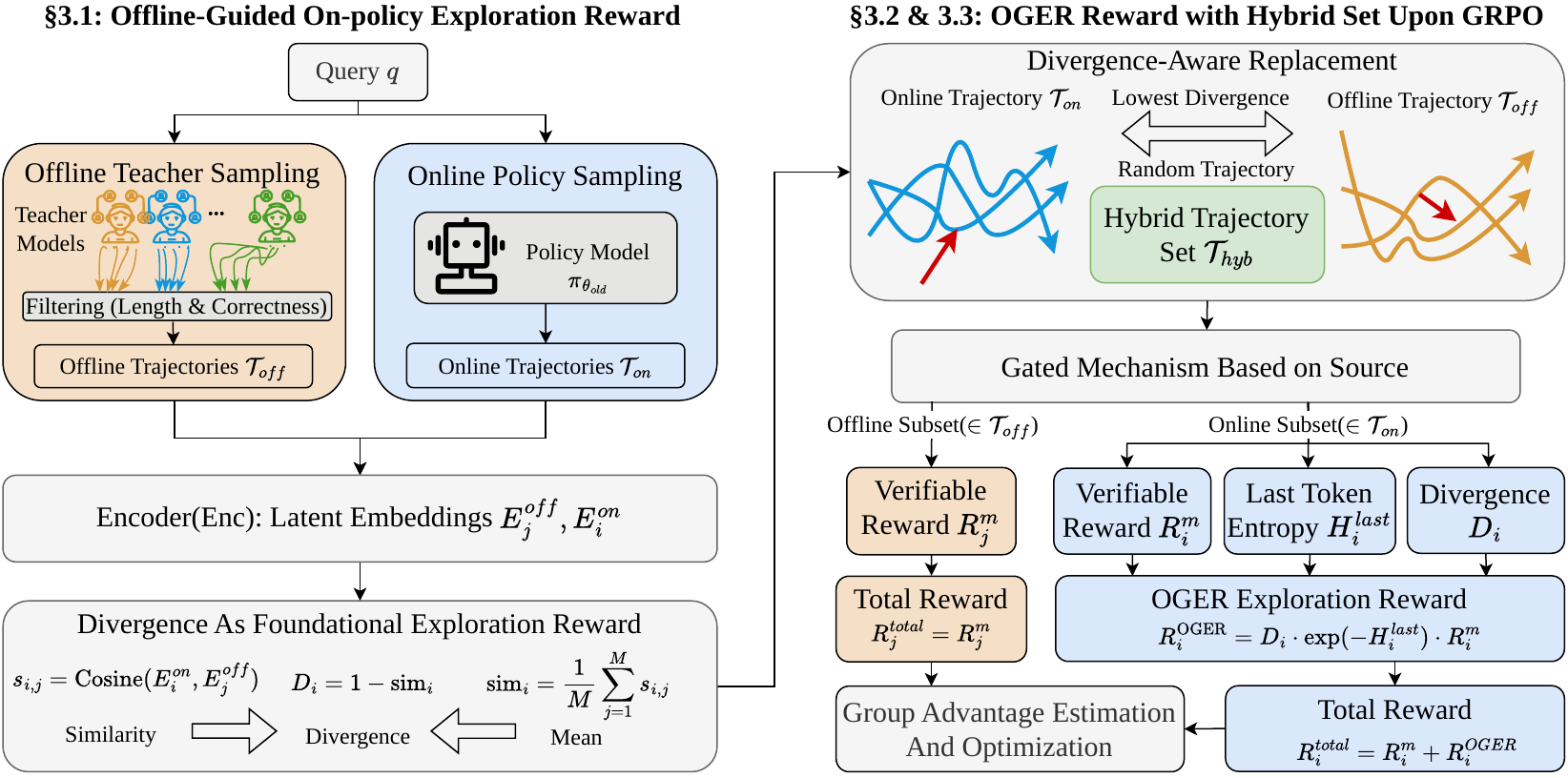}
    \caption{The overall architecture of the OGER framework. We first construct a comprehensive, high-quality offline dataset from trajectories generated by multiple teacher models and perform hybrid offline-online integration by mapping online sampled trajectories and offline references into a shared representation space to compute divergence. This divergence serves as the basis for our exploration reward and offline replacement. Furthermore, we leverage the Shannon entropy of the last token to refine the exploration reward, providing a fine-grained, uncertainty-aware modulation of the training signal.}
    \label{fig:main}
\end{figure*}

\section{OGER: Offline-Guided Exploration Reward for Hybrid RL}

Current research underscores the critical need for autonomous exploration to enhance LLM reasoning, particularly through hybrid training paradigms and entropy-based regularization. In this work, we introduce OGER, a framework that fosters synergistic integration of offline expert guidance and online exploratory discovery. Our approach departs from traditional data-mixing strategies by shifting the fusion of loss to the reward modeling level. 
Specifically, OGER constructs an auxiliary exploration reward that benchmarks online trajectories against an ensemble of high-quality multi-teacher offline trajectories by divergence. This provides the model with an explicit directional signal to navigate the complex reasoning search space. To ensure exploration is both diverse and well-calibrated, this guidance is further refined by a fine-grained mechanism that uses the model’s internal entropy of the distributions at the last token of the trajectory to shape the exploration reward. By unifying these dimensions, OGER effectively mitigates the risk of entropy collapse while transcending the limitations of passive imitation, thereby facilitating the discovery of novel reasoning paths. The comprehensive architecture of our method is illustrated in~\Cref{fig:main}.

\subsection{Offline-Guided On-policy Exploration Reward}
\label{sec:dataset}

\paragraph{Ensembling Offline Teacher Demonstrations}

A core component of the OGER framework is the construction of a high-quality dataset synthesized from multiple state-of-the-art teacher models. By generating supplementary trajectories across multiple teacher models, we capture a rich manifold of reasoning structures, particularly the long-form Chain-of-Thought (CoT) patterns that characterize advanced systemic reasoning. This multi-source synthesis yields a robust ensemble of standard trajectories, providing a high-fidelity offline reference for our reward-level fusion.

To ensure both data integrity and computational tractability, we implement two rigorous filtering constraints: (1) we apply a length-consistency filter to ensure all teacher trajectories align with a uniform maximum reasoning length, and (2) we employ a correctness filter for each trajectory, retaining only those with a correct final answer for training.

\paragraph{Embedding Trajectories for Comparison}
During the training procedure, for a given query $q$, let $\mathcal{T}_{on} = \{\tau^{on}_1, \dots, \tau^{on}_N\}$ denote a set of $N$ trajectories generated by the online policy model, and $\mathcal{T}_{off} = \{\tau^{off}_1, \dots, \tau^{off}_M\}$ represent $M$ offline trajectories curated from teacher models. 
Rather than treating offline trajectories as static replacements for online batches, we utilize them as a semantic reference to steer the online exploration process. We first embed both $\mathcal{T}_{off}$ and $\mathcal{T}_{on}$ into a shared $d$-dimensional latent space. For the $i$-th trajectory in $\mathcal{T}_{on}$ and the $j$-th trajectory in $\mathcal{T}_{off}$, the latent representations are defined as:
\begin{equation}
\rev{E^{on}_{i} = \text{Enc}(\tau^{on}_i), E^{off}_{j} = \text{Enc}(\tau^{off}_j) \in \mathbb{R}^{d}}
\label{eq:embedding}
\end{equation}

\paragraph{Exploration Reward for On-policy Trajectories}

In standard RLVR frameworks, the verifiable reward function is typically defined as follows: for a reward function $R(x, y) \in [0, 1]$, $x$ denotes the input query and $y$ represents the response generated by policy model $\pi_{\theta}$. Current research predominantly assigns rewards via binary classification based on final answer correctness~\citep{zhan2025exgrpolearningreasonexperience,zheng2025groupsequencepolicyoptimization}. The primary objective of RLVR training is to maximize the expected reward, formally expressed as:
\begin{align}
\mathcal{L}_{\text{RLVR}}(\theta) = \mathbb{E}_{x \sim \mathcal{D},\, y \sim \pi_\theta(\cdot \mid x)}[\, R(x, y) \,]
\end{align}
where $\theta$ denotes the model parameters and $\mathcal{D}$ represents the training distribution. 

However, existing methods typically assign uniform rewards to all correct trajectories, which inadvertently restricts the model's ability to selectively learn from a diverse manifold of reasoning paths. To address this limitation, we construct an exploration reward that distinguishes reward signals across different correct trajectories. 

Building upon the latent representations $E^{on}_i$ and $E^{off}_j$ defined in~\Cref{eq:embedding}, we then compute the cosine similarity between the online and offline trajectories:
\begin{equation}
\begin{aligned}
s_{i,j} &= \text{Cosine}(E^{on}_i,E^{off}_j)
\\&=\frac{E^{on}_i \cdot E^{off}_j}{\|E^{on}_i\|_2 \|E^{off}_j\|_2}
\end{aligned}
\end{equation}

To quantify the proximity of each online sample to the offline distribution, we calculate the mean similarity for each online trajectory:
\begin{equation}
\text{sim}_i = \frac{1}{M} \sum_{j=1}^{M} s_{i,j} 
\end{equation}
This similarity score reflects the degree to which the online model replicates the teacher's reasoning patterns; a higher similarity score indicates that the model has effectively converged toward the offline distribution through imitation. To incentivize the model to venture beyond these established patterns and explore a broader reasoning manifold, we utilize semantic divergence as the foundational exploration reward $D_i$ for trajectory $\tau^{on}_i$:
\begin{equation}
D_{i} = 1 - \text{sim}_{i}
\label{eq:rsim}
\end{equation}
\begin{table*}[!tbp]
    \centering
    \small
    \resizebox{\textwidth}{!}{
        \begin{tabular}{l l l l l l l l l }
            \toprule
            \rowcolor{tableheader}
            \best{Method} & \best{AIME 2024} & \best{AIME 2025} & \best{AMC} & \best{MATH-500} & \best{Minerva} & \best{Olympiad} & \best{OOD} & \best{AVG.} \\
            \midrule
            \rowcolor{groupheader}
            \multicolumn{9}{c}{\textbf{Qwen2.5-Math-1.5B}} \\
            \rowcolor{baselinebg}
            \textbf{Base} & 4.17 \deltafont{deltaneg}{-5.21} & 0.94 \deltafont{deltaneg}{-6.25} & 22.29 \deltafont{deltaneg}{-19.39} & 29.80 \deltafont{deltaneg}{-40.00} & 5.51 \deltafont{deltaneg}{-20.96} & 13.63 \deltafont{deltaneg}{-17.33} & 21.32 \deltafont{deltapos}{+5.95} & 13.95 \deltafont{deltaneg}{-14.74} \\
            \rowcolor{baselinebg}
            \textbf{SFT} & 4.48 \deltafont{deltaneg}{-4.90} & 7.81 \deltafont{deltapos}{+0.62} & 24.32 \deltafont{deltaneg}{-17.36} & 72.20 \deltafont{deltapos}{+2.40} & 19.12 \deltafont{deltaneg}{-7.35} & 33.04 \deltafont{deltapos}{+2.08} & 35.94 \deltafont{deltapos}{+20.57} & 28.13 \deltafont{deltaneg}{-0.56} \\
            \rowcolor{baselinebg}
            \textbf{GRPO} & 9.38 \deltafont{deltaneu}{0.00} & 7.19 \deltafont{deltaneu}{0.00}  & 41.68 \deltafont{deltaneu}{0.00}  & 69.80 \deltafont{deltaneu}{0.00}  & 26.47 \deltafont{deltaneu}{0.00}  & 30.96  \deltafont{deltaneu}{0.00}  & 15.37 \deltafont{deltaneu}{0.00}  & 28.69 \deltafont{deltaneu}{0.00}  \\
            \rowcolor{baselinebg}
            \textbf{Luffy} & \best{14.37} \deltafont{deltapos}{+4.99} & \second{9.69} \deltafont{deltapos}{+2.50} & \best{47.33} \deltafont{deltapos}{+5.65} & \second{75.20} \deltafont{deltapos}{+5.40} & 27.21 \deltafont{deltapos}{+0.74} & \best{39.85} \deltafont{deltapos}{+8.89} & 33.07 \deltafont{deltapos}{+17.70} & \second{35.25} \deltafont{deltapos}{+6.56} \\
            \rowcolor{baselinebg}
            \textbf{ExGRPO$^\star$} & 11.15 \deltafont{deltapos}{+1.77} & 9.38 \deltafont{deltapos}{+2.19} & 43.45 \deltafont{deltapos}{+1.77} & 72.80 \deltafont{deltapos}{+3.00} & \best{30.88} \deltafont{deltapos}{+4.41} & 31.85 \deltafont{deltapos}{+0.89} & \second{36.44} \deltafont{deltapos}{+21.07} & 33.71 \deltafont{deltapos}{+5.02} \\
            \rowcolor{block2bg}
            \textbf{OGER} & \second{14.27} \deltafont{deltapos}{+4.89} & \best{13.54} \deltafont{deltapos}{+6.35} & \second{46.76} \deltafont{deltapos}{+5.08} & \best{76.60} \deltafont{deltapos}{+6.80} & \second{29.41} \deltafont{deltapos}{+2.94} & \second{39.11} \deltafont{deltapos}{+8.15} & \best{37.70} \deltafont{deltapos}{+22.33} & \best{36.77} \deltafont{deltapos}{+8.08} \\
            \hdashline

            \rowcolor{groupheader}
            \multicolumn{9}{c}{\textbf{Qwen2.5-Math-7B}} \\
            \rowcolor{baselinebg}
            \textbf{Base} & 13.65 \deltafont{deltaneg}{-3.75} & 5.63 \deltafont{deltaneg}{-5.52} & 44.92 \deltafont{deltaneg}{-9.82} & 65.00 \deltafont{deltaneg}{-15.20} & 14.71 \deltafont{deltaneg}{-25.00} & 30.07 \deltafont{deltaneg}{-9.34} & 26.28 \deltafont{deltaneg}{-4.93} & 28.61 \deltafont{deltaneg}{-10.51} \\
            \rowcolor{baselinebg}
            \textbf{SFT$^\star$} & 25.52 \deltafont{deltapos}{+8.12} & \second{23.02} \deltafont{deltapos}{+11.87} & 58.21 \deltafont{deltapos}{+3.47} & 82.80 \deltafont{deltapos}{+2.60} & 41.18 \deltafont{deltapos}{+1.47} & 43.26 \deltafont{deltapos}{+3.85} & 50.16 \deltafont{deltapos}{+18.95} & 46.31 \deltafont{deltapos}{+7.19} \\
            \rowcolor{baselinebg}
            \textbf{GRPO} & 17.40 \deltafont{deltaneu}{0.00}  & 11.15 \deltafont{deltaneu}{0.00}  & 54.74 \deltafont{deltaneu}{0.00}  & 80.20 \deltafont{deltaneu}{0.00}  & 39.71 \deltafont{deltaneu}{0.00}  & 39.41 \deltafont{deltaneu}{0.00}  & 31.21 \deltafont{deltaneu}{0.00}  & 39.12 \deltafont{deltaneu}{0.00}  \\
            \rowcolor{baselinebg}
            \textbf{Luffy} & 26.67 \deltafont{deltapos}{+9.27} & 21.04 \deltafont{deltapos}{+9.89} & 63.59 \deltafont{deltapos}{+8.85} & \second{87.00} \deltafont{deltapos}{+6.80} & \second{42.28} \deltafont{deltapos}{+2.57} & \second{48.74} \deltafont{deltapos}{+9.33} & 51.33 \deltafont{deltapos}{+20.12} & \second{48.66} \deltafont{deltapos}{+9.54} \\
            \rowcolor{baselinebg}
            \textbf{ExGRPO$^\star$} & \second{29.90} \deltafont{deltapos}{+12.50} & 18.23 \deltafont{deltapos}{+7.08} & \second{64.01} \deltafont{deltapos}{+9.27} & 83.80 \deltafont{deltapos}{+3.60} & 40.07 \deltafont{deltapos}{+0.36} & 45.63 \deltafont{deltapos}{+6.22} & \best{55.35} \deltafont{deltapos}{+24.14} & 48.14 \deltafont{deltapos}{+9.02} \\
            \rowcolor{block2bg}
            \textbf{OGER} & \best{31.77} \deltafont{deltapos}{+14.37} & \best{25.10} \deltafont{deltapos}{+13.95} & \best{68.64} \deltafont{deltapos}{+13.90} & \best{88.40} \deltafont{deltapos}{+8.20} & \best{45.22} \deltafont{deltapos}{+5.51} & \best{53.48} \deltafont{deltapos}{+14.07} & \second{51.61} \deltafont{deltapos}{+20.40} & \best{52.03} \deltafont{deltapos}{+12.91} \\
            \bottomrule
        \end{tabular}
    }
    \caption{Comprehensive performance comparison of OGER against other methods using the \textbf{Qwen2.5-Math-1.5B} and \textbf{Qwen2.5-Math-7B} backbones. 
    Values in parentheses denote the absolute performance gain or loss relative to the \textbf{GRPO} baseline. We highlight the \textbf{best} and \underline{second-best} results for each model scale using bold and underlined text, respectively. \rev{$^\star$ denotes results we reproduce by evaluating the open-source checkpoints released by the corresponding methods.}
}
\label{tab:main}
\end{table*}
\subsection{OGER Reward with Hybrid Set}
\paragraph{Entropy-aware Reward Refinement}
In RLVR, rewards are sparse and tied solely to final-answer correctness, offering no signal about the reliability of the reasoning process and thus risking reward hacking~\citep{skalse2022defining}. \rev{Our foundational reward $D_i$ inherits this risk, as it indiscriminately rewards any divergence from the teacher, including erratic or unreliable deviations. To calibrate it, we draw on prior findings that the predictive distribution at the final token reflects the model's culminating judgment after the reasoning chain has converged, with its entropy capturing the associated aleatoric uncertainty~\citep{yang2025laserreinforcementlearninglasttoken}. We thus use the Shannon entropy of the last token in $\tau^{on}_i$ as a confidence proxy to refine $D_i$ from~\Cref{eq:rsim}, amplifying confident divergence while suppressing erratic exploration.}

Let $\mathcal{V}$ denote the vocabulary of the policy model and $p(v)$ represent the probability distribution over tokens $v \in \mathcal{V}$. The entropy of the distribution at the final token position of trajectory $\tau^{on}_i$ is defined as:
\begin{equation}
    H_{i}^{last} = -\sum_{v \in \mathcal{V}} p(v) \log p(v)
\label{eq:entropy}
\end{equation}

By integrating the foundational exploration reward $D_{i}$ with the last-token entropy $H_{i}^{last}$, the refined exploration reward for trajectory $\tau^{on}_i$ is formulated as:
\begin{equation}
    \rev{R_{i}^{\text{OGER}} = D_{i} \cdot \exp(-H_{i}^{last}) \cdot R^m_i} 
\label{eq:roger}
\end{equation}
where $R_i^{m} \in \{0, 1\}$ denotes the standard verifiable reward. This formulation ensures that $R^{\text{OGER}}_i$ modulates the reward exclusively for correct online trajectories.

\paragraph{Divergence-Aware Trajectory Replacement}
\label{sec:off-policy-training}

In the joint training procedure, for each problem query $q$, we integrate the online and offline distributions by replacing the trajectory in $\mathcal{T}_{on}$ that exhibits the lowest divergence with a randomly sampled trajectory from $\mathcal{T}_{off}$. 
This results in a hybrid trajectory set $\mathcal{T}_{hyb}$ comprising $N$ samples. 

Notably, we apply the proposed exploration reward exclusively to on-policy trajectories. For offline teacher trajectories, we utilize only the standard verifiable reward $R^{m}$. This categorical separation ensures that exploration signals—which are intrinsically coupled to the model's internal uncertainty—do not introduce noise into the standard teacher demonstrations.

The final reward used to compute group advantages is defined as a gated composition of verifiable and exploratory signals. For any trajectory $\tau_i \in \mathcal{T}_{hyb}$, the total reward $R_i^{total}$ is formulated as:
\begin{equation}
R_{i}^{total} = 
\begin{cases} 
R_{i}^{m}  + R_{i}^{\text{OGER}}, & \text{if } \tau_i \in \mathcal{T}_{on} \\
R_{i}^m, & \text{if } \tau_i \in \mathcal{T}_{off}
\end{cases}
\label{eq:rtotal}
\end{equation}
In summary, within the OGER framework, the auxiliary exploration reward is exclusively assigned to correct trajectories generated by the online policy. For offline teacher demonstrations and incorrect online samples, the reward remains aligned with the standard verifiable baseline.

\subsection{Implementation with GRPO}

In the GRPO framework, $\pi_{\theta_{\text{old}}}$ denotes the policy model used for trajectory sampling, while $\pi_{\theta}$ represents the policy undergoing optimization. Each trajectory in $\mathcal{T}_{hyb}$ is evaluated via \Cref{eq:rtotal} to further determine its relative advantage within the group. For a specific trajectory $\tau_{i} \in \mathcal{T}_{hyb}$, the group-relative advantage $A_{i}$ is computed as:
\begin{equation}
A_{i} = \frac{R_i^{\text{total}} - \text{mean}(\{R_j^{\text{total}}|j=1,\cdots,N\})}{\text{std}(\{R_j^{\text{total}}|j=1,\cdots,N\})}
\label{eq:advantage}
\end{equation}

The GRPO objective function for optimizing the parameters $\theta$ is formulated as:
\begin{equation}
\begin{aligned}
\mathcal{J}_{\text{GRPO}}(\theta) &= \frac{1}{\sum_{i=1}^{N} |\tau_i|} \sum_{i=1}^{N} \sum_{t=1}^{|\tau_i|} \mathrm{CLIP}(r_{i,t}(\theta), A_i, \epsilon) 
\\ &- \beta \cdot \mathbb{D}_{\mathrm{KL}}[\pi_\theta || \pi_{\mathrm{ref}}]
\end{aligned}
\end{equation}
where $r_{i,t}(\theta) = \pi_{\theta}(\tau_{i,t} \mid q, \tau_{i,<t}) /\pi_{\theta_{\text{old}}}(\tau_{i,t} \mid q, \tau_{i,<t})$ represents the importance sampling ratio. We follow the advantage-shaping methodology proposed by \citet{luffy} to optimize the importance sampling term. 

The Kullback-Leibler (KL) divergence term $\mathbb{D}_{\text{KL}}[\pi_\theta || \pi_{\text{ref}}]$, scaled by the hyperparameter $\beta$, traditionally regularizes the policy against a reference model $\pi_{\text{ref}}$ to mitigate catastrophic forgetting or distribution collapse. However, consistent with recent advancements, which suggest that the KL penalty may become redundant when employing robust advantage normalization and intrinsic exploration rewards~\citep{hu2025open,yu2025dapo,cui2025process}, we omit the KL term in our primary training objective.

\section{Experiments}
\subsection{Datasets}
\paragraph{Training Dataset} 
\rev{We utilize a subset of the \textbf{OpenR1-Math-220k}~\footnote{https://huggingface.co/datasets/open-r1/OpenR1-Math-220k} corpus, curated and filtered by~\citet{luffy}, as our foundational dataset. It comprises 45k instances, each paired with a \textbf{DeepSeek-R1} reasoning trajectory for tasks sourced from NuminaMath 1.5~\citep{numina_math_datasets}. Building upon this, we construct a heterogeneous multi-teacher ensemble (\Cref{sec:dataset}) by incorporating additional trajectories distilled from \textbf{Qwen3-32B}~\citep{yang2025qwen3} and \textbf{GLM-4.5 Air}~\citep{5team2025glm45agenticreasoningcoding}; the curation procedure and dataset statistics are detailed in~\Cref{apd:detail_data_filter}. While OGER leverages the full multi-teacher ensemble, we restrict the Luffy baseline~\citep{luffy} to DeepSeek-R1 samples only, ensuring strict adherence to its original configuration for a fair comparison. We further examine the effect of data diversity in~\Cref{sec:diversity}}.

\paragraph{Evaluation Benchmarks}
We evaluate across six mathematical reasoning benchmarks: \textbf{AIME 2024}, \textbf{AIME 2025}, \textbf{AMC}~\citep{li2024numinamath}, \textbf{Minerva}~\citep{lewkowycz2022solving}, \textbf{OlympiadBench}~\citep{he2024olympiadbench}, and \textbf{MATH-500}~\citep{hendrycks2021measuring}; and three out-of-domain (\textbf{OOD}) tasks: \textbf{ARC-Challenge}~\citep{clark2018think}, \textbf{GPQA-Diamond}~\citep{rein2024gpqa}, and \textbf{MMLU-Pro}~\citep{wang2024mmlu} to assess robustness and generalization. We report pass@1 throughout\rev{, and for the OOD tasks we report the average across the three datasets}; the full evaluation protocol is provided in~\Cref{apd:detail_data_eval}.

\begin{table*}[!tbp]
    \centering
    \setlength{\aboverulesep}{0pt} \setlength{\belowrulesep}{0pt}
    \small
    \resizebox{\textwidth}{!}{
        \begin{tabular}{l l l l l l l l l}
            \toprule
            \rowcolor{tableheader}
            \best{Method} & \best{AIME 2024} & \best{AIME 2025} & \best{AMC} &\best{MATH-500} & \best{Minerva} & \best{Olympiad} & \best{OOD} & \best{AVG.} \\
            \midrule
            \rowcolor{groupheader}
            \multicolumn{9}{c}{\textbf{OGER Variants}} \\
            \rowcolor{baselinebg}
            \textbf{OGER w/o Refinement} & 30.31 \deltafont{deltaneg}{-1.46} & 23.75 \deltafont{deltaneg}{-1.35} & 66.42 \deltafont{deltaneg}{-2.22} & 88.40 \deltafont{deltaneu}{0.00} & 41.18 \deltafont{deltaneg}{-4.04} & 53.04 \deltafont{deltaneg}{-0.44} & 53.22 \deltafont{deltapos}{+1.61} & 50.90 \deltafont{deltaneg}{-1.13} \\
            \rowcolor{baselinebg}
            \textbf{OGER w/o Reward} & 27.08 \deltafont{deltaneg}{-4.69} & 23.13 \deltafont{deltaneg}{-1.97} & 66.64 \deltafont{deltaneg}{-2.00} & 88.40 \deltafont{deltaneu}{0.00} & 36.03 \deltafont{deltaneg}{-9.19} & 50.81 \deltafont{deltaneg}{-2.67} & 51.19 \deltafont{deltaneg}{-0.41} & 49.04 \deltafont{deltaneg}{-2.99} \\
            \hdashline
            \rowcolor{groupheader}
            \multicolumn{9}{c}{\textbf{OGER with Different Replace Density}} \\
            \rowcolor{baselinebg}
            \textbf{OGER w/ 2-Replace} & 27.92 \deltafont{deltaneg}{-3.85} & 25.21 \deltafont{deltapos}{+0.11} & 67.09 \deltafont{deltaneg}{-1.55} & 87.40 \deltafont{deltaneg}{-1.00} & 38.60 \deltafont{deltaneg}{-6.62} & 52.89 \deltafont{deltaneg}{-0.59} & 51.66 \deltafont{deltapos}{+0.05} & 50.11 \deltafont{deltaneg}{-1.92} \\
            \rowcolor{baselinebg}
            \textbf{OGER w/ 3-Replace} & 28.75 \deltafont{deltaneg}{-3.02} & 24.58 \deltafont{deltaneg}{-0.52} & 64.65 \deltafont{deltaneg}{-3.99} & 88.20 \deltafont{deltaneg}{-0.20} & 38.60 \deltafont{deltaneg}{-6.62} & 51.11 \deltafont{deltaneg}{-2.37} & 49.72 \deltafont{deltaneg}{-1.88} & 49.37 \deltafont{deltaneg}{-2.66} \\
            \bottomrule
        \end{tabular}
    }
    \caption{Ablation studies of the OGER framework on mathematical reasoning benchmarks. \textbf{OGER w/o Refinement} and \textbf{OGER w/o Reward} denote variants where the entropy-aware refinement and the exploration reward are, respectively, removed. We further evaluate OGER under varying replacement densities (\textbf{OGER w/ 2-Replace} and \textbf{OGER w/ 3-Replace}). Values in parentheses denote the absolute performance gain or loss relative to the \textbf{OGER}.}
    \label{tab:ablation study}
\end{table*}

\subsection{Implementation Details}

All primary experiments use \textbf{Qwen2.5-Math-7B} and \textbf{Qwen2.5-Math-1.5B}~\citep{yang2024qwen25mathtechnicalreportmathematical}. We omit the KL penalty and adopt the token-level advantage shaping of~\citet{luffy}. For each query we sample $N=8$ online trajectories and apply the hybrid update in~\Cref{sec:off-policy-training}, replacing one online trajectory with a high-quality offline teacher trajectory. Trajectory correctness is verified by Math-Verify~\footnote{https://github.com/huggingface/Math-Verify}, and embeddings are computed with bge-large-en-v1.5~\citep{bge_embedding}.

\paragraph{Baselines}
To rigorously evaluate the efficacy of OGER, we compare it against the following baselines: (1) \textbf{Base}: the zero-shot performance of the pre-trained backbone without any fine-tuning, serving as the performance floor. \rev{(2) \textbf{SFT}: supervised fine-tuning of the backbone directly on the offline teacher trajectories, representing pure imitation without any reward signal.} (3) \textbf{GRPO}: the standard on-policy paradigm that applies the GRPO algorithm~\citep{shao2024deepseekmath} to our primary dataset, representing a purely online reinforcement learning approach. (4) \textbf{Luffy}~\citep{luffy}: a competitive hybrid baseline that incorporates DeepSeek-R1 trajectories during training with token-level advantage shaping. \rev{(5) \textbf{ExGRPO}~\citep{zhan2025exgrpolearningreasonexperience}: a hybrid method that further augments online training with an experience-replay mechanism, reusing valuable historical trajectories to improve efficiency.}
Full training configurations and system prompt are reported in~\Cref{apd:exp_detail}.

\subsection{Main Results}
The primary experimental results are summarized in~\Cref{tab:main}. Overall, OGER achieves substantial gains across both model scales, reaching \textbf{36.77} on the 1.5B model and \textbf{52.03} on the 7B model. This corresponds to relative improvements of \textbf{28.2\%} and \textbf{33.0\%} over the vanilla GRPO baseline, and absolute gains of \textbf{1.52} and \textbf{3.37} points over the stronger Luffy baseline, respectively, confirming the scalability of OGER from small- to large-scale models. \rev{Moreover, OGER consistently surpasses both the pure imitation baseline (SFT) and the competitive offline--online hybrid method ExGRPO, demonstrating the effectiveness of our reward-level fusion over alternative integration strategies.}

The advantages of OGER are most pronounced on challenging reasoning benchmarks, with the 7B variant outperforming all baselines across most evaluation tasks. On AIME 2024 and AIME 2025, OGER 7B reaches \textbf{31.77} and \textbf{25.10}, a clear leap over Luffy's 26.67 and 21.04; at the 1.5B scale, it attains \textbf{13.54} on AIME 2025, outperforming Luffy by nearly \textbf{40\%}. Crucially, these gains extend beyond mathematics: on OOD tasks, OGER reaches \textbf{37.70} at 1.5B and ranks second at 7B with \textbf{51.61}. This indicates that our entropy-refined exploration reward does not merely memorize teacher distributions but actively incentivizes more robust and transferable reasoning. Consistent improvements on specialized benchmarks such as MATH-500 and Minerva further validate the universality of our exploration mechanism.

\rev{Finally, since OGER is a reward design that is agnostic to the underlying RL algorithm, we further adapt it to the DAPO algorithm and train on Qwen2.5-Math-1.5B. As detailed in~\Cref{apd:dapo}, OGER delivers consistent gains beyond the GRPO setting, demonstrating its generality across different optimization paradigms.}

\section{Analysis}
\subsection{Ablation Studies}

\paragraph{Effectiveness of Reward Components}
To rigorously evaluate the contribution of each proposed component, we analyze two variants of our framework: (1) \textbf{OGER w/o Refinement}: this variant excludes the entropy-aware refinement mechanism, relying solely on the foundational exploration reward. (2) \textbf{OGER w/o Reward}: this configuration utilizes the standard verifiable reward and offline data integration without exploration reward, serving as a baseline to validate the necessity of the exploration reward.

As summarized in~\Cref{tab:ablation study}, removing the entropy-aware refinement results in consistent performance degradation across all benchmarks relative to the full OGER framework. Specifically, while OGER w/o Refinement achieves a 2.24 point improvement over the Luffy baseline, it still falls 1.13 points short of the full OGER model on average. Furthermore, when the exploration reward is entirely removed (OGER w/o Reward), and only offline trajectory replacement is used, the model yields only marginal gains over Luffy and remains significantly inferior to the full OGER configuration. 
These results underscore that while offline data provides a valuable semantic reference, the entropy-modulated exploration reward is the primary catalyst for superior reasoning performance and training stability. Crucially, our findings suggest that merely increasing the diversity of offline trajectories fails to yield substantial gains. This indicates that static imitation alone is insufficient and the dynamic interaction between offline alignment and entropy-driven discovery is essential.

\paragraph{Impact of Replacement Density and Diversity}
\label{sec:diversity}

\textbf{Density.} In our primary OGER configuration, we replace a single online trajectory per query group ($N=8$). To assess sensitivity to offline guidance intensity, we compare against \textbf{OGER w/ 2-Replace} and \textbf{OGER w/ 3-Replace}, which substitute two and three online trajectories per group, respectively. As shown in~\Cref{tab:ablation study}, although heavier replacement still surpasses the Luffy baseline, it exhibits a clear gap relative to the standard OGER. This indicates that the density of offline guidance must be carefully calibrated: excessive reliance on expert trajectories stifles the model's autonomous exploration, whereas single-sample replacement leaves enough room for OGER to incentivize novel reasoning beyond mere imitation of the teacher distribution.

\textbf{Diversity.} \rev{To disentangle the effect of teacher diversity from that of our reward design, we examine the \textbf{OGER w/o Reward} variant, which removes all OGER-specific rewards and thus reduces to the native Luffy objective trained on the full multi-teacher data. Compared with the original Luffy trained solely on DeepSeek-R1 trajectories, this variant yields only a marginal average gain (49.04 vs.\ 48.66). This shows that OGER's improvements do not stem from merely accessing additional external data; rather, OGER exploits this diversity to improve the correctness of the model's online sampling.}
\begin{figure*}[!htbp]
    \centering
    \begin{subfigure}[b]{0.33\linewidth}
        \centering
        \includegraphics[width=\linewidth]{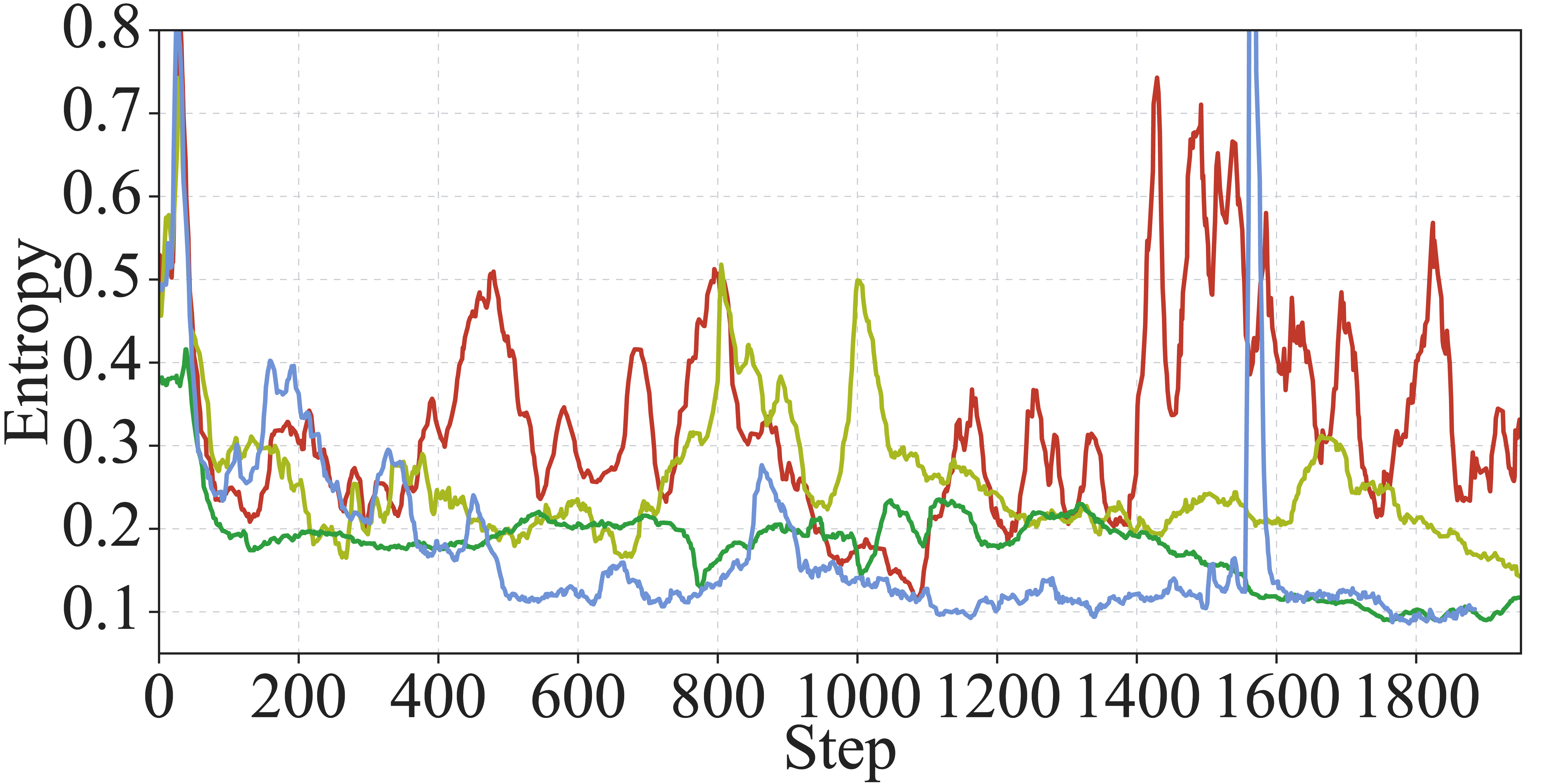} 
        \caption{Entropy}
        \label{fig:sub1}
    \end{subfigure}%
    \hfill 
    \begin{subfigure}[b]{0.33\linewidth}
        \centering
        \includegraphics[width=\linewidth]{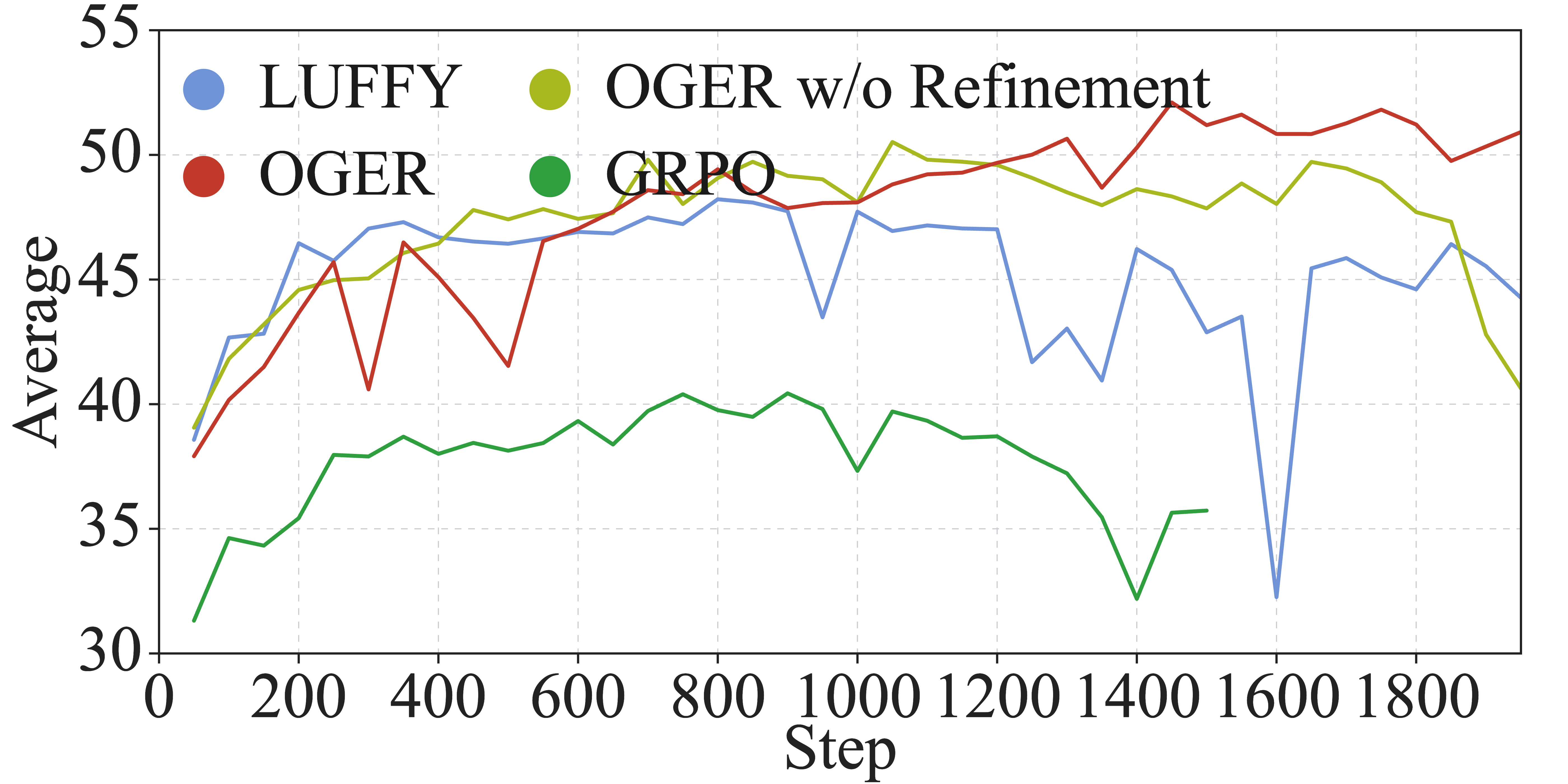}
        \caption{Average Score}
        \label{fig:sub2}
    \end{subfigure}%
    \hfill 
    \begin{subfigure}[b]{0.33\linewidth}
        \centering
        \includegraphics[width=\linewidth]{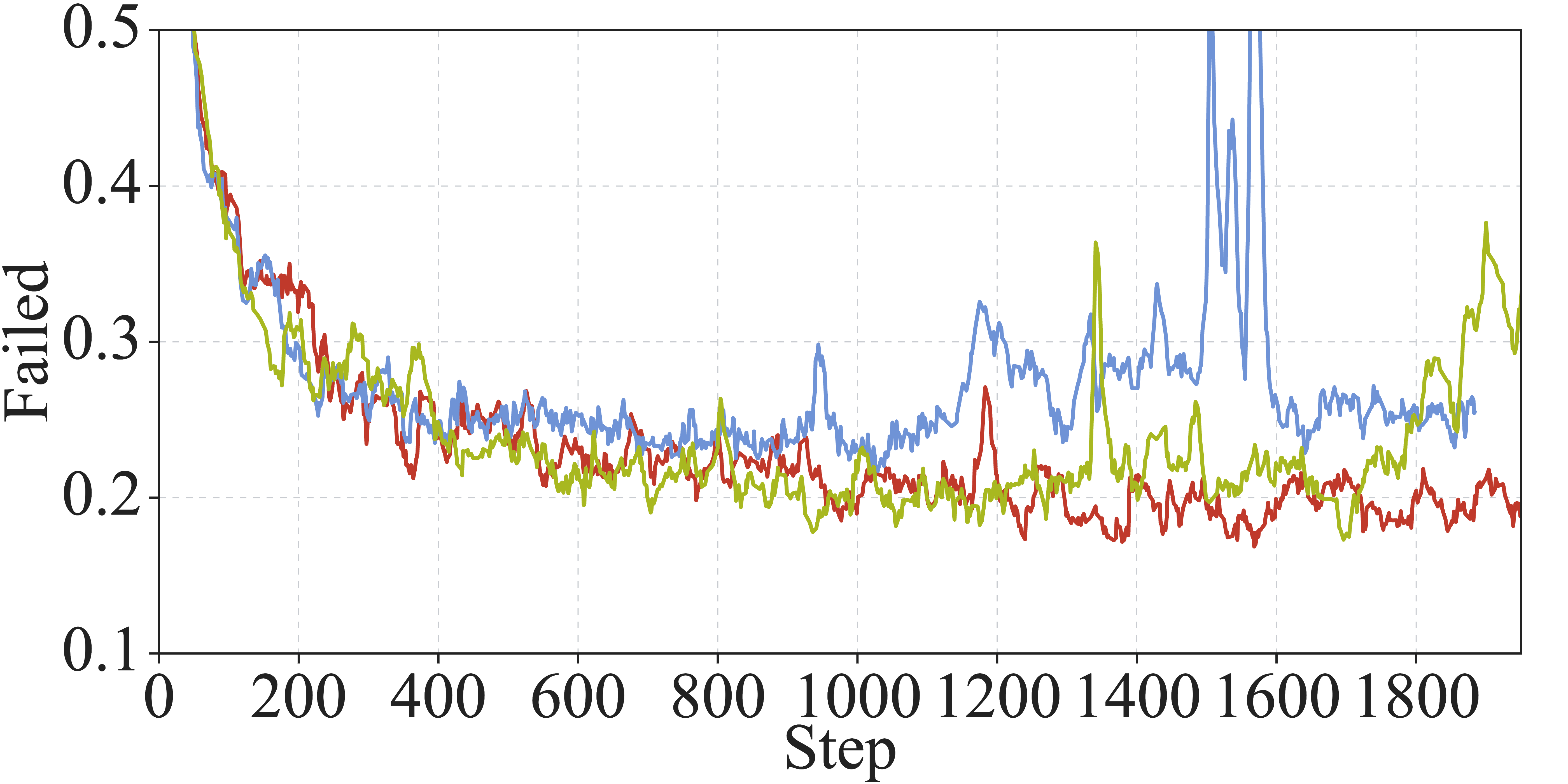}
        \caption{Failed Response Ratio}
        \label{fig:sub3}
    \end{subfigure}
    
    \caption{Comparative analysis of training dynamics across OGER, its variant OGER w/o Refinement, and baselines (GRPO and Luffy). \textbf{Left}: evolution of policy entropy throughout the training process. \textbf{Middle}: progression of average accuracy across six major mathematical benchmarks. \textbf{Right}: the failure ratio within training responses (GRPO's data is not available).}
    \label{fig:dynamic}
\end{figure*}

\subsection{Training Dynamics}
To gain deeper insights into the underlying optimization process, we provide an extensive analysis of the OGER framework utilizing the Qwen2.5-Math-7B model. Specifically, we examine the evolution of policy entropy, mean test accuracy, and the proportion of failed responses in the training batch. These metrics are visualized in~\Cref{fig:sub1}, \Cref{fig:sub2}, and \Cref{fig:sub3}, respectively. The training resource consumption is detailed in~\Cref{tab:training ultilization}. 

As illustrated in~\Cref{fig:sub1}, both OGER and its variant OGER w/o Refinement maintain significantly higher entropy levels throughout the training lifecycle compared to Luffy and GRPO. While standard on-policy RL often suffers from rapid entropy collapse—where the policy prematurely concentrates on a narrow set of high-reward trajectories—our framework effectively preserves a broader exploratory landscape. This sustained entropy correlates with the performance trends in~\Cref{fig:sub2}, where OGER demonstrates superior training stability. Unlike the baselines, which exhibit performance plateaus or degradation as the policy becomes overly confident, OGER facilitates longer optimization cycles by consistently discovering novel reasoning paths. The detailed analysis of our OGER reward during training, as provided in~\Cref{apd:Balance}, further demonstrates that our method maintains continuous exploration throughout the learning process.
\begin{table}[t]
    \centering
    \small 
    \begin{tabular}{lrrr}
    \toprule
    \multirow{2}{*}{\textbf{Method}} & \multirow{2}{*}{\textbf{GPU Hours}} & \multicolumn{2}{c}{\textbf{Data Usage}} \\
     &  & \textbf{Online} & \textbf{Offline} \\ \midrule
    \textbf{GRPO} & 75*8 & 45K*8 & 0K \\
    \textbf{Luffy} & 120*8 & 45K*7 & 45K \\
    \textbf{OGER} & 168*8 & 45K*7 & 128K \\ 
    \bottomrule
    \end{tabular}
    \caption{Comparison of resource requirements between OGER and other methods.}
    \label{tab:training ultilization}
\end{table}
Furthermore,~\Cref{fig:sub3} shows that our method exhibits a markedly lower failure rate as training progresses. This decline suggests that OGER effectively navigates the reasoning manifold, translating exploratory signals into tangible improvements in problem-solving coverage. 

Regarding computational overhead, we summarize resource utilization in~\Cref{tab:training ultilization}. While OGER requires additional training time per step due to exploration reward computations and the overhead of maintaining higher sample diversity, this investment directly translates into substantial gains in final model accuracy and robustness. The higher compute cost is thus justified by the model's ability to transcend the imitation bottleneck and achieve better performance.
\begin{figure}[ht]
    \centering
    \includegraphics[width=0.8\linewidth]{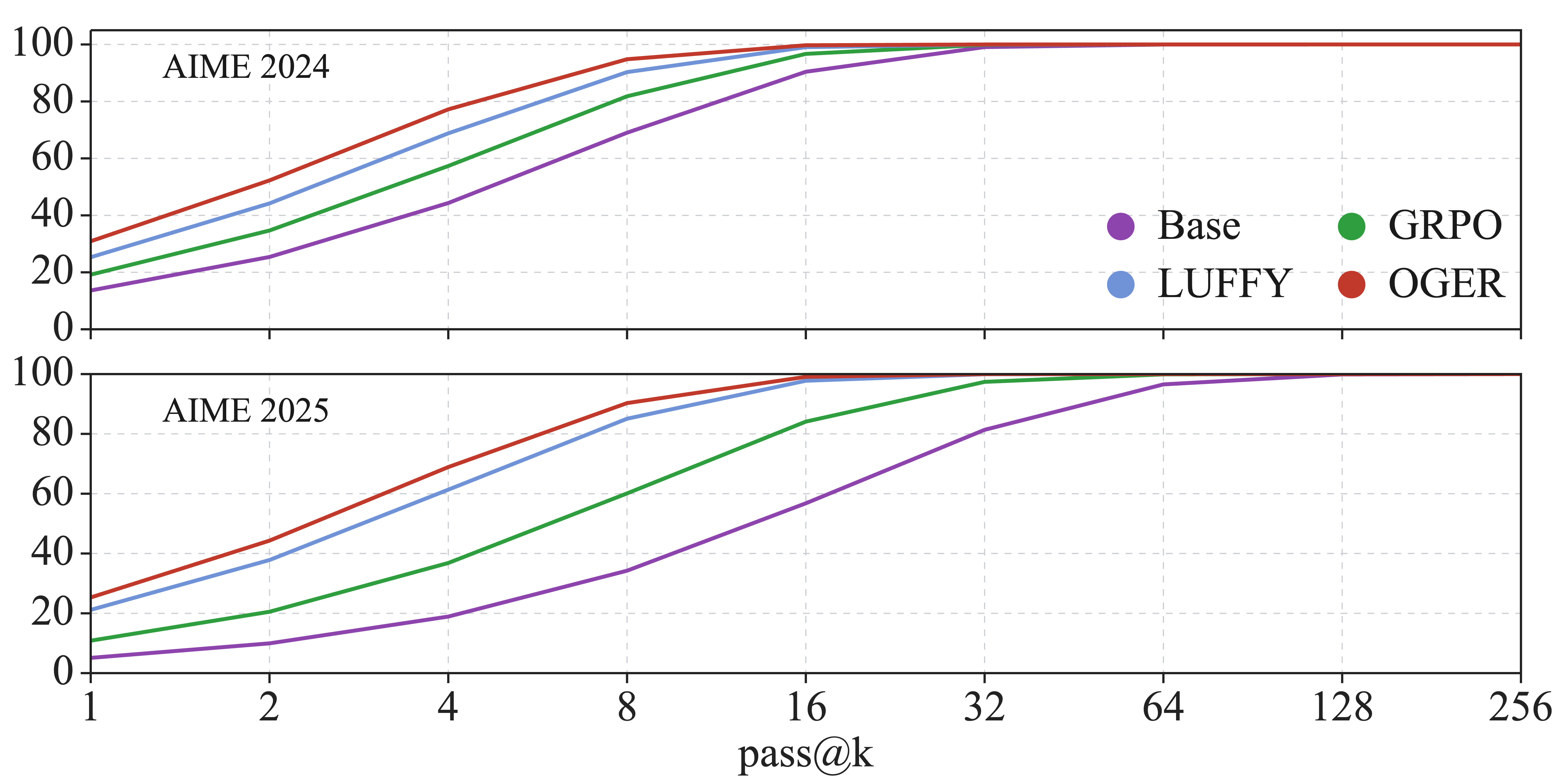}
    
    \caption{pass@k performance on AIME 2024 and AIME 2025 using 256 rollouts. Our proposed OGER method consistently outperforms all baselines across various k values, demonstrating a significantly higher convergence rate in solvability coverage.}
    \label{fig:coverage}
\end{figure}

\subsection{Coverage of the Reasoning Manifold}
To evaluate the breadth of the reasoning space the model has mastered, we assess pass@k performance for OGER and the selected baselines on the AIME 2024 and AIME 2025 benchmarks. As illustrated in~\Cref{fig:coverage}, we generated 256 rollout samples per problem to compare solvability coverage across varying computational budgets. Our proposed OGER framework, along with its variants, consistently outperforms both Luffy and GRPO across the entire range of k. 

These results demonstrate that our entropy-aware reward refinement not only enhances the precision of individual reasoning trajectories but also significantly expands the aggregate coverage of solvable problems. By maintaining a high-entropy exploratory regime, OGER ensures that the model navigates a more diverse and robust solution manifold. This broader coverage indicates that our method successfully facilitates the discovery of alternative reasoning paths without compromising the model's fundamental generalization capacity.

Furthermore, we conducted extensive experiments across a range of decoding temperatures to evaluate the model's inference-time robustness. The details can be found in~\Cref{apd:test_time}. OGER consistently outperforms Luffy and GRPO across the entire temperature spectrum, which substantiates the effectiveness of our approach in enhancing the model's intrinsic reasoning capabilities. 

\section{Conclusion}
We propose \textbf{OGER}, a simple yet powerful hybrid training framework that integrates offline guidance with online entropy to build auxiliary exploration rewards. Experimental results across multiple benchmarks demonstrate that OGER significantly enhances mathematical reasoning performance and exhibits superior generalization to out-of-domain tasks, achieving \textbf{4.3\%} and \textbf{6.9\%} improvements on average for both small- and large-scale models compared to a competitive baseline. These findings underscore the feasibility and immense potential of entropy, alongside offline demonstrations, to guide model capability growth. 

\section*{Limitations}
\rev{While OGER delivers consistent gains across model scales and RL algorithms, it has a few limitations. First, computing the exploration reward requires embedding online and offline trajectories during training, introducing moderate additional overhead over standard on-policy RL. Second, OGER builds on external teacher trajectories and a sentence-embedding model, and uses the last-token entropy as a lightweight confidence proxy; exploring richer teacher sources, embedding choices, and uncertainty estimates is a promising direction. Finally, our experiments focus on mathematical reasoning with Qwen2.5-Math backbones; extending OGER to larger models and broader domains is left for future work.}



\bibliography{custom}
\appendix
\section{The Details of Training Data and Evaluation}
\label{apd:detail_data}
\subsection{Dataset Filtering}
\label{apd:detail_data_filter}
\rev{We construct the multi-teacher offline dataset by sampling reasoning trajectories from several high-quality open-source models, namely DeepSeek-R1, Qwen3-32B, and GLM-4.5 Air.} We retain only trajectories that pass two constraints: a correctness filter, which keeps trajectories whose final answer is verified by Math-Verify, and a length-consistency filter with a maximum sequence length of 8k tokens.

\begin{figure}[!tp]
    \centering
    \includegraphics[width=0.8\linewidth]{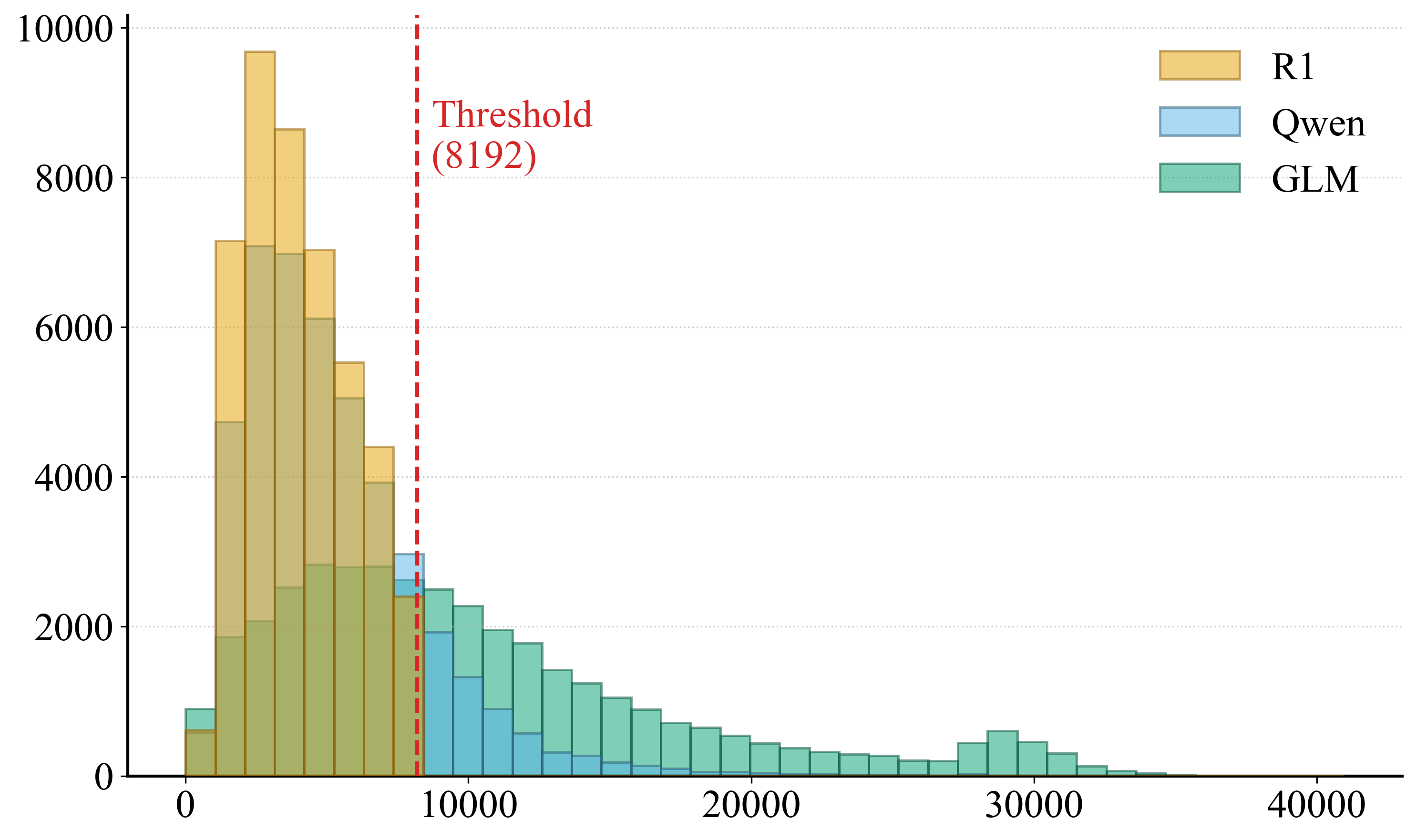}
    \caption{Distribution of sequence lengths for trajectories generated by multiple teacher models. We filter out samples with more than 8k tokens. \textbf{R1} denotes DeepSeek-R1, \textbf{Qwen} means Qwen3-32B, and \textbf{GLM} represents GLM-4.5 Air.}
    \label{fig:offline distribution}
\end{figure}

\rev{~\Cref{tab:filtered datasets} and~\Cref{fig:offline distribution} summarize the statistics of the curated dataset. The three teacher models exhibit distinct generation profiles: DeepSeek-R1 contributes the most valid samples with the highest accuracy and the shortest average length, whereas GLM-4.5 Air produces substantially longer trajectories at a lower pass rate. This heterogeneity supplies the diverse reasoning patterns that OGER leverages as offline references.}

\begin{table}[!tp]
    \centering
    \small 
    \begin{tabular}{lccc}

        \toprule

        \textbf{Model} & \textbf{Valid Samples} & \textbf{Avg. Length} & \textbf{Accuracy (\%)} \\

        \midrule

        \textbf{R1}            & 45,462 & 4,021.14            & 99.28        \\

        \textbf{Qwen}         & 36,958 & 5,252.09            & 94.90                 \\

        \textbf{GLM}        & 17,887 & 10,318.07           & 82.14                 \\

        \bottomrule

    \end{tabular}
    \caption{Statistics of the curated offline dataset generated by multiple teacher models. The \textbf{Valid Samples} refer to trajectories that passed the correctness verification and length constraints.}
    \label{tab:filtered datasets}
\end{table}

\begin{table*}[!tbp]
    \centering
    \setlength{\aboverulesep}{0pt} \setlength{\belowrulesep}{0pt}
    \small
    \resizebox{\textwidth}{!}{
        \begin{tabular}{l l l l l l l l l }
            \toprule
            \rowcolor{tableheader}
            \best{Method} & \best{AIME 2024} & \best{AIME 2025} & \best{AMC} & \best{MATH-500} & \best{Minerva} & \best{Olympiad} & \best{OOD} & \best{AVG.} \\
            \midrule
            \rowcolor{baselinebg}
            \textbf{DAPO} & \second{14.58} \deltafont{deltaneu}{0.00} & \second{11.46} \deltafont{deltaneu}{0.00} & \second{47.59} \deltafont{deltaneu}{0.00} & \best{77.40} \deltafont{deltaneu}{0.00} & 30.88 \deltafont{deltaneu}{0.00} & 36.30 \deltafont{deltaneu}{0.00} & \second{35.94} \deltafont{deltaneu}{0.00} & \second{36.31} \deltafont{deltaneu}{0.00} \\
            \rowcolor{baselinebg}
            \textbf{OGER w/o Reward} & 13.44 \deltafont{deltaneg}{-1.14} & \best{11.56} \deltafont{deltapos}{+0.10} & 47.03 \deltafont{deltaneg}{-0.56} & 76.60 \deltafont{deltaneg}{-0.80} & \second{31.62} \deltafont{deltapos}{+0.74} & \second{36.59} \deltafont{deltapos}{+0.29} & 34.04 \deltafont{deltaneg}{-1.91} & 35.84 \deltafont{deltaneg}{-0.47} \\
            \rowcolor{block2bg}
            \textbf{OGER} & \best{14.90} \deltafont{deltapos}{+0.32} & 10.73 \deltafont{deltaneg}{-0.73} & \best{48.00} \deltafont{deltapos}{+0.41} & \second{77.20} \deltafont{deltaneg}{-0.20} & \best{32.72} \deltafont{deltapos}{+1.84} & \best{40.30} \deltafont{deltapos}{+4.00} & \best{37.34} \deltafont{deltapos}{+1.39} & \best{37.31} \deltafont{deltapos}{+1.00} \\
            \bottomrule
        \end{tabular}
    }
    \caption{\rev{Performance of OGER instantiated with the DAPO algorithm on the Qwen2.5-Math-1.5B backbone. \textbf{DAPO} denotes the standard DAPO algorithm; \textbf{OGER w/o Reward} retains only offline trajectory replacement on the multi-teacher data without the exploration reward; and \textbf{OGER} is our full method. OGER consistently improves over both baselines, confirming that its benefits transfer beyond GRPO to other RL algorithms.}}
    \label{tab:dapo}
\end{table*}

\subsection{Evaluation Protocol}
\label{apd:detail_data_eval}
For benchmarks with limited test samples (AIME 2024, AIME 2025, and AMC), we report pass@1 at 32 rollouts to mitigate variance and better capture reasoning consistency; for the remaining benchmarks we report pass@1 at a single rollout. For OOD tasks, we report the average pass@1 (\textbf{OOD}) at a single rollout. All tasks are evaluated at a sampling temperature of $0.8$ to balance generation diversity and precision.

\section{Experiments Detail}
\label{apd:exp_detail}
\paragraph{Training Configurations}
The Qwen2.5-Math backbone has an original context length of 4,096 tokens. To accommodate complex reasoning and align with the offline teacher trajectory lengths, we extend the context window to 16,384 tokens, accordingly adjusting the RoPE theta from 10,000 to 40,000. For RL, we train with a learning rate of $1\times10^{-6}$, a global batch size of $128$, and a micro-batch size of $64$, fixing the entropy loss coefficient at $0.01$ and omitting the KL divergence penalty. For each query, we sample $N=8$ online trajectories at a temperature of $1.0$, and the hybrid update (\Cref{sec:off-policy-training}) replaces one online trajectory per query with a high-quality offline trajectory from the teacher ensemble. We train the Qwen2.5-Math-7B model for up to 1,950 steps and the Qwen2.5-Math-1.5B variant for 1,500 steps. All RL experiments are conducted on a single cluster of 8$\times$ NVIDIA H200 GPUs with a tensor parallelism (TP) degree of 2.
\rev{For the SFT baseline, we set the learning rate to $1\times10^{-5}$ and train the Qwen2.5-Math-1.5B model for 2,000 steps. For Qwen2.5-Math-7B, we directly adopt the SFT checkpoint released by Luffy~\citep{luffy}, which was trained under the same configuration as ours.}

\begin{table*}[htbp]
    \centering
    \begin{tabularx}{0.95\linewidth}{>{\raggedright\arraybackslash}X}
        \toprule
        \textbf{System Prompt} \\
        \midrule
        Your task is to follow a systematic, thorough reasoning process before providing the final solution. This involves analyzing, summarizing, exploring, reassessing, and refining your thought process through multiple iterations. Structure your response into two sections: Thought and Solution. In the Thought section, present your reasoning using the format: \texttt{"<think>\textbackslash n \{thoughts\} </think>\textbackslash n"}. Each thought should include detailed analysis, brainstorming, verification, and refinement of ideas. After \texttt{"</think>\textbackslash n"}, in the Solution section, provide the final, logical, and accurate answer, clearly derived from the exploration in the Thought section. If applicable, include the answer in \texttt{\textbackslash boxed\{\}} for closed-form results like multiple choices or mathematical solutions. \\
        \textbf{User:} This is the problem: \{Question\} \\
        \textbf{Assistant:} \texttt{<think>} \\
        \bottomrule
    \end{tabularx}
    \caption{The system prompt used for all experiments.}
    \label{tab:system_prompt}
\end{table*}
\paragraph{System Prompt}
For all experiments, we use the same system prompt in~\Cref{tab:system_prompt} for training. At evaluation, we keep it for the mathematical tasks and \rev{remove it for the OOD tasks}.

\section{Extending OGER to the DAPO Algorithm}
\label{apd:dapo}
\rev{The OGER reward design is agnostic to the underlying RL algorithm. To further demonstrate its generality across different optimization paradigms, we apply OGER to the DAPO algorithm and conduct experiments on Qwen2.5-Math-1.5B, using the same experimental setup as our GRPO training, with one exception: since standard DAPO struggles to generate valid training data containing the \texttt{<think>} tag on the base model without offline data, we remove the \texttt{<think>} component from the system prompt during vanilla DAPO training. As shown in~\Cref{tab:dapo}, OGER remains effective under DAPO: it improves over both vanilla DAPO and the OGER w/o Reward variant (i.e., Luffy trained on the full multi-teacher data) by \textbf{1.00} and \textbf{1.47} points, respectively. These results confirm the effectiveness of OGER across multiple RL algorithms.}

\section{The Balance of Imitation and Exploration}
\label{apd:Balance}
\begin{figure}[t]
\centering
\includegraphics[width=0.8\linewidth]{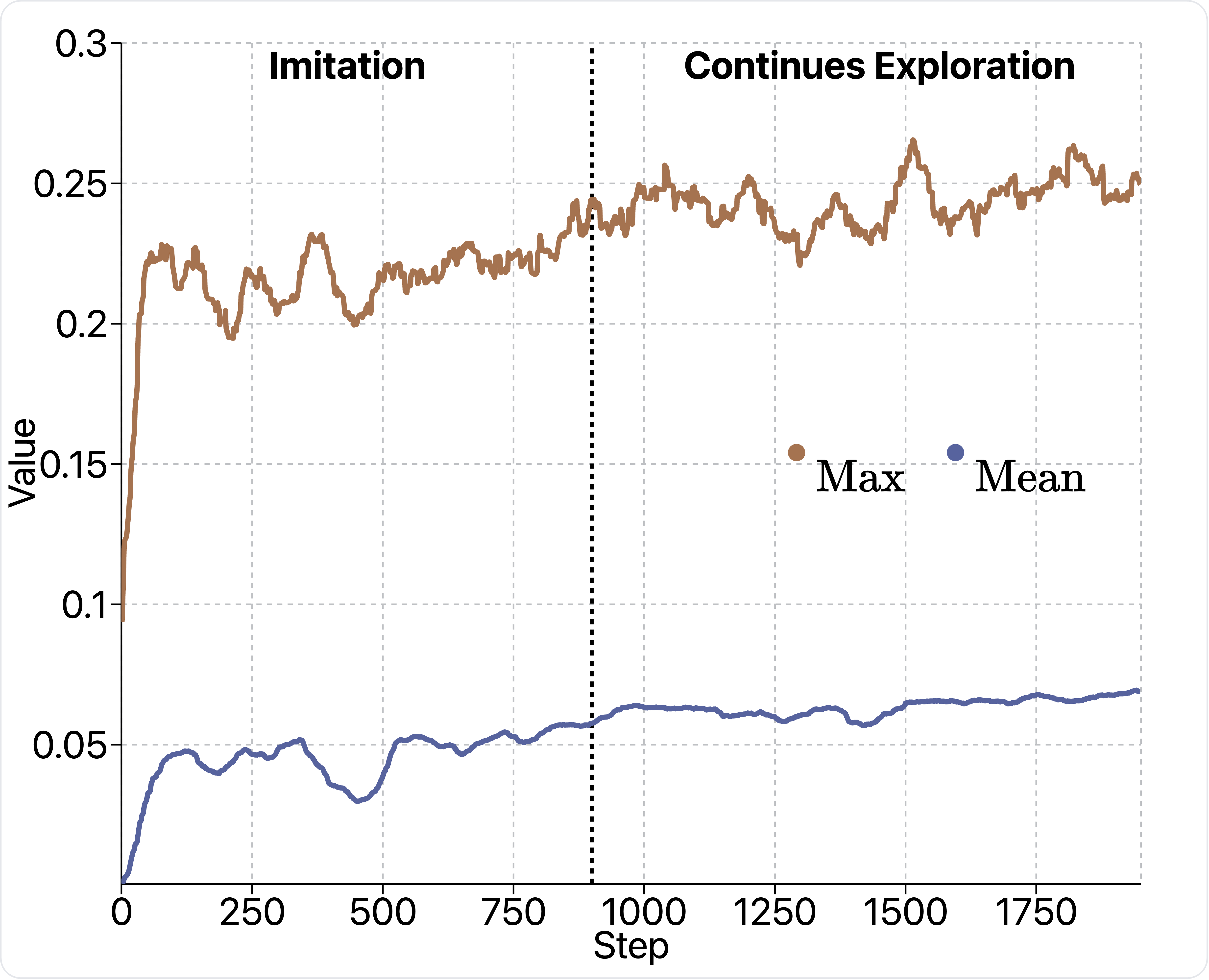}
\caption{The evolution of the OGER exploration reward during training. The transition from low to stabilized reward levels reflects the shift from imitation to continuous exploration.}
\label{fig:oger-reward}
\end{figure}

To further dissect the behavioral shift of the model during the training process, we analyze the evolution of the OGER reward (both Mean and Max values), as illustrated in~\Cref{fig:oger-reward}. 

Our exploration reward is formulated based on the semantic divergence between online and offline trajectories. In the \textbf{early stages} of training, due to the model's limited initial reasoning capacity, the generation of correct online samples is relatively scarce. Consequently, the optimization focus primarily gravitates toward internalizing the high-quality reasoning patterns from the offline teacher trajectories. As training progresses into the \textbf{mid-to-late stages}, the model's intrinsic reasoning proficiency matures, leading to an increase in exploratory signals. The OGER reward subsequently stabilizes at a specific plateau, facilitating a dual-objective learning process: it continuously absorbs the sophisticated reasoning logic of the teacher models while simultaneously leveraging the exploration reward to autonomously discover novel reasoning paths that deviate from the static offline demonstrations. This dynamic transition effectively balances expert-guided imitation with autonomous latent space exploration.

\begin{figure}[t]
\centering
\includegraphics[width=0.8\linewidth]{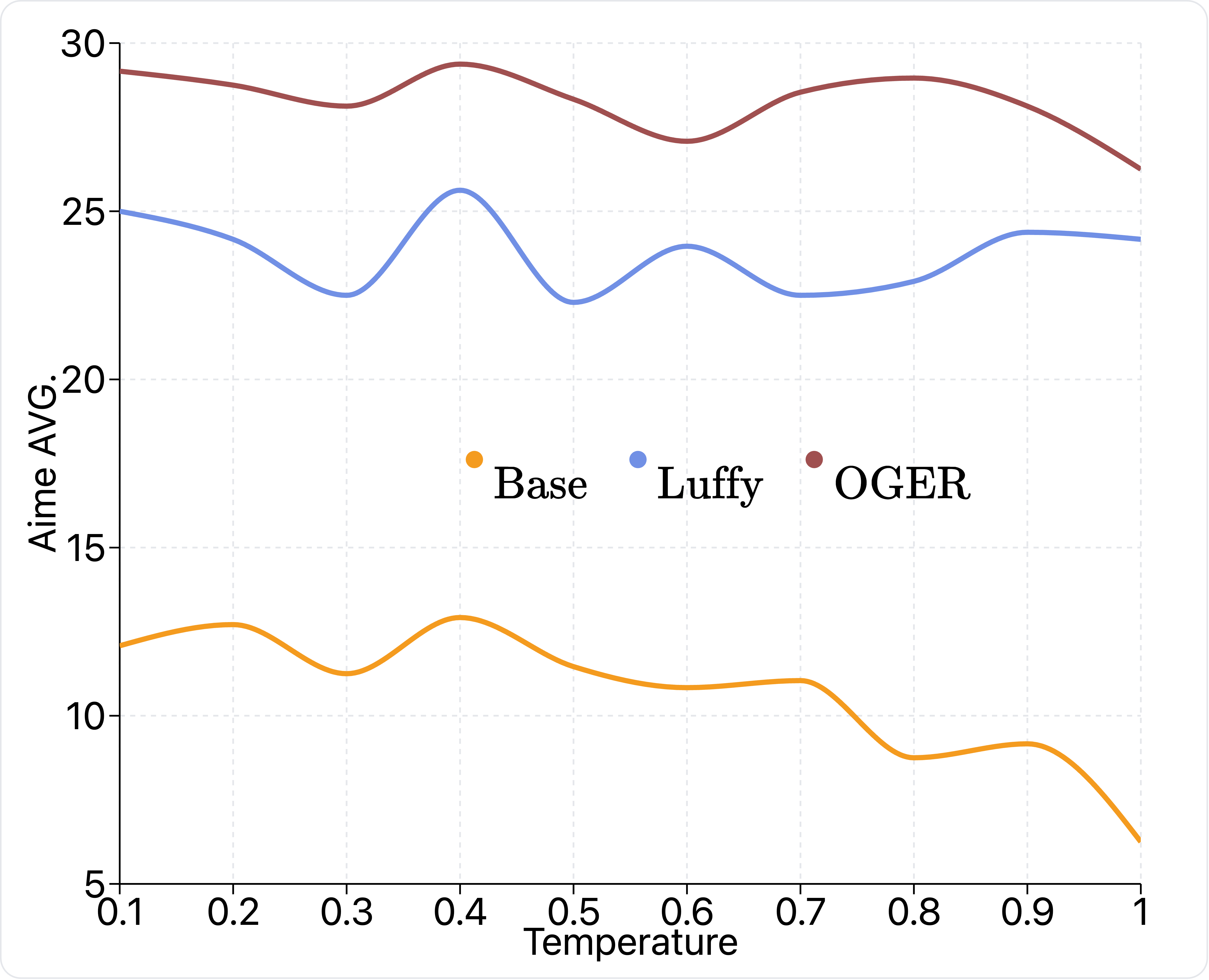}
\caption{The pass@8 performance across different inference temperatures on AIME 2024 and AIME 2025, we illustrate the average score.}
\label{fig:multi-tmp}
\end{figure}
\section{OGER Maintains Exploration During Inference}
\label{apd:test_time}
To evaluate the robustness of our framework across varying degrees of exploration during inference, we compare the average pass@8 scores of OGER, Luffy, and GRPO on the AIME 2024 and AIME 2025 benchmarks. We systematically vary the decoding temperature from $0.1$ to $1.0$ in increments of $0.1$.

As illustrated in~\Cref{fig:multi-tmp}, our proposed OGER framework consistently and significantly outperforms both Luffy and GRPO across the entire temperature spectrum. This consistency highlights OGER's stability under diverse exploratory constraints at test time. Furthermore, a horizontal comparison reveals that OGER maintains a relatively stable performance profile as the temperature increases. In contrast, GRPO exhibits a marked performance degradation at higher temperatures. This phenomenon underscores OGER's enhanced generalizable reasoning capability; whereas baseline models may become erratic or lose reasoning coherence when stochasticity increases, OGER preserves a structured and effective reasoning manifold, effectively bridging the gap between training-time exploration and test-time robustness.

\end{document}